\documentclass[letterpaper, 10 pt, conference]{ieeeconf}  

\IEEEoverridecommandlockouts                              

\makeatletter
\let\NAT@parse\undefined
\makeatother

\usepackage{graphics} 
\usepackage{epsfig} 
\usepackage{amsmath} 
\usepackage{amssymb}  
\usepackage{bbm}
\usepackage{dsfont}
\usepackage{algorithm}
\usepackage{algorithmicx}
\usepackage[noend]{algpseudocode}
\usepackage{array}
\usepackage[font=small]{caption}
\usepackage{subcaption}
\usepackage{booktabs}
\usepackage{tabularx}
\usepackage{verbatim}

\usepackage[colorlinks,bookmarks=true,citecolor=blue,urlcolor=blue,linktocpage=true]{hyperref}
\usepackage[numbers,sort&compress]{natbib}

\usepackage{bibentry}

\usepackage{setspace}
\AtBeginDocument{%
  \addtolength\abovedisplayskip{-0.2\baselineskip}%
  \addtolength\belowdisplayskip{-0.2\baselineskip}%
}

\algnewcommand\algorithmicinput{\textbf{Inputs:}}
\algnewcommand\algorithmicoutput{\textbf{Output:}}
\algnewcommand\INPUT{\item[\algorithmicinput]}
\algnewcommand\OUTPUT{\item[\algorithmicoutput]}

\newtheorem{theorem}{Theorem}

\newcommand\numberthis{\addtocounter{equation}{1}\tag{\theequation}}

\title{\LARGE \bf
PERCH: Perception via Search for\\ Multi-Object Recognition and Localization}

\author{Venkatraman Narayanan \and Maxim Likhachev
\thanks{The Robotics Institute, Carnegie Mellon University,
                Pittsburgh, USA
               {\tt\small \{venkatraman,maxim\} at cs.cmu.edu}.
							                 }%
}

\begin{document}

\maketitle
\thispagestyle{empty}
\pagestyle{empty}

\begin{abstract}
  In many robotic domains such as flexible automated manufacturing or personal
  assistance, a fundamental perception task is that of identifying and
  localizing objects whose 3D models are known. Canonical approaches to this
  problem include discriminative methods that find correspondences between
  feature descriptors computed over the model and observed data. While these
  methods have been employed successfully, they can be unreliable when the
  feature descriptors fail to capture variations in observed data; a classic
  cause being occlusion. As a step towards deliberative reasoning, we
  present PERCH: PErception via SeaRCH, an algorithm that seeks to find the
  best explanation of the observed sensor data by hypothesizing possible
  scenes in a generative fashion. Our contributions are: i) formulating
  the multi-object recognition and localization task as an optimization
  problem over the space of hypothesized scenes, ii) exploiting structure in
  the optimization to cast it as a combinatorial search problem on what we
  call the Monotone Scene Generation Tree, and iii) leveraging
  parallelization and recent advances in multi-heuristic search in
  making combinatorial search tractable.  We prove that our system can
  guaranteedly produce the best explanation of the scene under the chosen
  cost function, and validate our claims on real world RGB-D test data. Our
  experimental results show that we can identify and localize objects under
  heavy occlusion---cases where state-of-the-art methods struggle.
\end{abstract}

\section{Introduction}
\label{sec:introduction}

A ubiquitous robot perception task is that of identifying and localizing
objects whose 3D models are known ahead of time: examples include robots operating in
flexible automation factory settings, and domestic robots manipulating common household objects.
Traditional methods for identifying and localizing objects rely on a
two-step procedure:
i) precompute a set of feature descriptors on the 3D models and match them
to observed features, and ii) estimate the rigid transform between the set of
found correspondences. In more recent methods, global
descriptors jointly encoding object pose and viewpoint information are computed
over different training instances, and a lookup is performed at test time.
While such discriminative methods have been used successfully, they are
limited by the ability of the feature descriptors to capture variations in
observed data. For illustration, consider a scene with two objects, one almost
completely occluding the other. Methods that employ feature correspondence
matching fare poorly as key feature descriptors could be lost
due to occlusion (Fig.~\ref{fig:chess}), whereas learning-based methods could suffer as they might
have not seen a similar training instance where the object is only partially
visible.

We introduce an orthogonal approach to tackle this problem: Perception via Search
(PERCH), which exploits the fact that the full 6 DoF sensor pose is available for most robotic
systems. PERCH is a generative approach that attempts to simulate or render
the scene which best explains the observed data. Our hypothesis is that if 3D
models and sensor pose are available, we could perform deliberative
reasoning: ``under this configuration of objects in the scene for the given
sensor pose, we would expect to see only this specific portion of that
object". The ability to reason deliberatively paves the way for an exhaustive
search over possible configurations of objects. 

While exhaustive search provides optimal solutions, it is often impractical
owing to the size of the state space that grows exponentially with the number
of objects in the scene. A key insight in this work is that the exhaustive
search over possible scene configurations can be formulated as a tree search
problem for a specific choice of an `explanation cost'. The formulation
involves breaking down the scene explanation cost into additive components
over individual objects in the scene, which in turn manifest as edge costs in
a tree called the Monotone Scene Generation Tree.  This allows us to
use state-of-the-art heuristic search techniques for determining the
configuration of objects that best explains the observed scene. We summarize
our contributions below:


\begin{figure}[t!]
\centering
\begin{minipage}{.5\columnwidth}
  \vspace*{\fill}
  \centering
  \includegraphics[width=0.97\columnwidth, height=3cm]{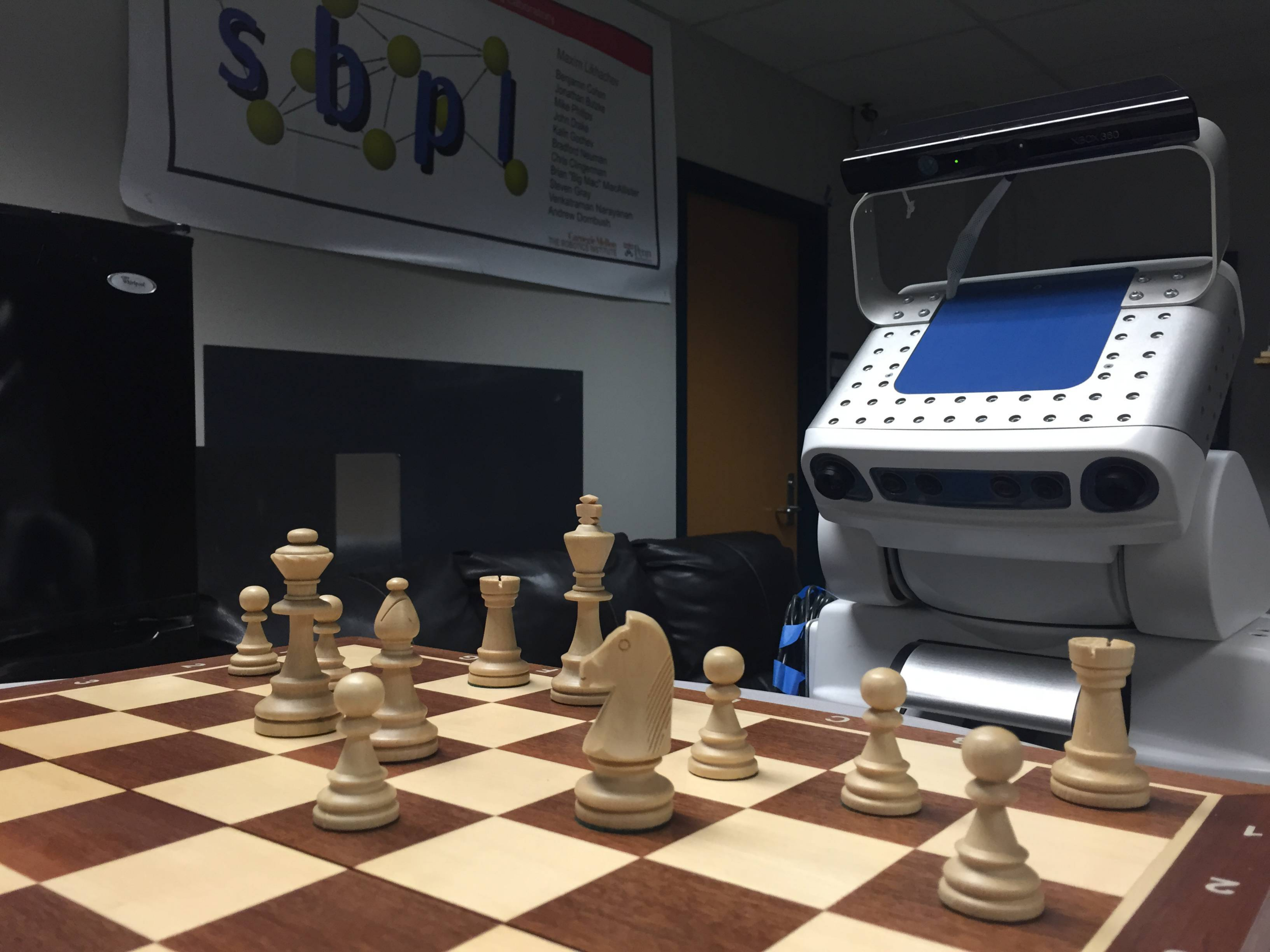}
  \subcaption{}
  \label{chess_real}
\end{minipage}%
\begin{minipage}{.5\columnwidth}
  \vspace*{\fill}
  \centering
  \includegraphics[width=0.97\columnwidth, height=1.435cm]{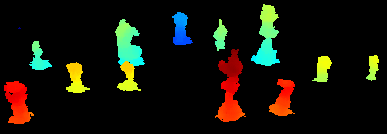}

	\vspace{1.3mm}
  \includegraphics[width=0.97\columnwidth, height=1.435cm]{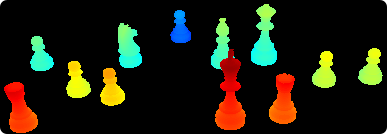}
  \subcaption{}
  \label{chess_depth}
\end{minipage}
\caption{Identifying and localizing the pose of multiple objects simultaneously is
  challenging due to inter-object occlusions. \subref{chess_real} A scene showing multiple chess pieces occluding
each other.
  \subref{chess_depth} \emph{Top:} The depth image from a Kinect sensor, colored by range. \emph{Bottom:} The best-match depth image produced by our algorithm PERCH through searching over possible poses of the chess pieces.}
  \label{fig:chess}
	\vspace{-4mm}
\end{figure}

\begin{itemize}
    \item Perception via Search (PERCH), an algorithm for
      simultaneously localizing multiple objects in 2.5D or 3D sensor data when 3D
      models of those objects are available along with the camera pose.
    \item Formulating the multi-object localization problem as the
      minimization of an `explanation cost' that captures the difference between
      the observed scene and the hypothesized scene.
      \item Exploiting structure in the explanation cost to cast it as a
        combinatorial search problem on a tree which we call the Monotone Scene Generation Tree.
      This alleviates the need to exhaustively generate/synthesize every
      possible scene while still returning solutions are that are provably optimal or bounded
      suboptimal.
    \item Incorporating parallelism in the search
      procedure, thereby allowing the algorithm to scale with the availability
      of computation.
\end{itemize}

In our experiments, we show how PERCH can localize objects even under
heavy occlusion---a result which would be hard to obtain without explicit
deliberative reasoning.

\section{Related Work}
\label{sec:related_work}
While model-based recognition and pose estimation of objects has been an
active area of research for decades in the computer vision
community~\cite{roberts1963machine,brooks1981symbolic, lowe1987three}, the
proliferation of low-cost depth sensors such as the Microsoft Kinect has
introduced a plethora of opportunities and challenges. We describe approaches
in vogue for object recognition and localization from 3D sensor data, their
limitations, inspirations from early research in vision that motivate our
work, and the potential role of contemporary learning-based systems.

\subsection{Local and Global 3D Feature Descriptors}
Model-based object recognition and localization in the present 3D era falls
broadly under two approaches: \emph{local} and \emph{global} recognition
systems. The former class of algorithms operate in a two step procedure: i)
compute and find correspondences between a set of local shape-preserving 3D
feature descriptors on the model and the observed scene and ii) estimate the
rigid transform between a set of geometrically consistent correspondences. A
final, optional and often used step is to 
perform a fine-grained local optimization to align the model to the scene and obtain
the pose. Examples of local 3D feature descriptors range from Spin
Images~\cite{johnson1999using} to Fast Point Feature Histograms
(FPFH)~\cite{rusu2009fast}, whereas final alignment procedures include
Iterative Closest Point (ICP)~\cite{chen1991object} and Bingham Procrustrean
Alignment (BPA)~\cite{glover2013bingham}. The survey paper by Aldoma et
al.~\cite{aldoma2012point} provides a comprehensive overview of other local
approaches.

The second, \emph{global} recognition systems employ a single-shot process for
identifying object type and pose jointly. Global feature descriptors encode the
notion of an object and capture shape and viewpoint information jointly in the
descriptor. These approaches employ a training phase to build a library of
global descriptors corresponding to different observed instances (e.g., each
object viewed from different viewpoints) and attempt to match the descriptor
computed at observation time to the closest one in the library. Additionally,
global methods unlike the local ones, require points in the observed scene to
be segmented into different clusters, so that descriptors can be computed on
each object cluster separately. Some of the global recognition systems include
Viewpoint Feature Histogram (VFH)~\cite{rusu2010fast}, Clustered Viewpoint
Feature Histogram (CVFH)~\cite{aldoma2011cad}, OUR-CVFH~\cite{aldoma2012our},
Ensemble of Shape Functions (ESF)~\cite{wohlkinger2011ensemble}, and Global
Radius-based Surface Descriptors (GRSD)~\cite{marton2011combined}.  Other
approaches to estimating object pose include local voting
schemes~\cite{drost2010model} or template
matching~\cite{hinterstoisser2013model} to first detect objects, and then using
global descriptor matching or ICP for pose refinement. 

Although both local and global feature-based approaches have enjoyed popularity
owing to their speed and intuitive appeal, they suffer when used for
identifying and localizing multiple objects in the scene. The limitation is
perhaps best described by the following lines from the book by Stevens and
Beveridge~\cite{stevens2000integrating}: ``\emph{Searching for individual objects in
isolation precludes explicit reasoning about occlusion. Although the absence of
a model feature can be detected (i.e., no corresponding data feature), the
absence cannot be explained (why is there no corresponding data feature?). As
the number of missing features increase, recognition performance degrades}''.
Global verification ~\cite{aldoma2012global, aldoma2013multimodal} and
filtering~\cite{pimentel2013multi} approaches somewhat attempt to address the occlusion
problems faced by feature-based methods through a joint optimization procedure
over candidate object poses, but are restricted by the fact
that initial predictions for object poses are provided by a system that does
not model occlusion.  In this work, we aim to explicity reason about the
interactions between multiple objects in the observed data by hypothesizing or
rendering scenes, and using combinatorial search to control the number of
scenes generated. 

\subsection{Search and Rendering-based Approaches}
The idea of using search to `explain' scenes was popular in the early years of
2D computer vision: Goad~\cite{goad1987special} promoted the idea of treating
feature matching between the observed scene and 3D model as a constrained
search while Lowe~\cite{lowe1987viewpoint} developed and implemented a
viewpoint-constrained feature matching system.
Grimson~\cite{grimson1987localizing} introduced the Interpretation Tree to
systematically search for geometrically-consistent matches between scene and
model features, while using various heuristics to speed up search. Our work is
also based on a search system, but it differs from the aforementioned works in
that the search is over the space of full hypothesized/rendered scenes and not
feature correspondences. In fact, our proposed algorithm does not employ
feature descriptors at all.

The philosophy of the Render, Match and Refine (RMR) approach proposed by
Stevens and Beveridge~\cite{stevens2000localized} motivates our work. RMR
explicitly models interaction between objects by rendering the
scene and uses occlusion data to inform measurement of similarity between the rendered and
observed scenes. It then uses a global optimization procedure to iteratively
improve the rendered scene to match the observed one. PERCH, our proposed
algorithm, operates on a similar philosophy but differs in several details.
The `explanation cost' we use to compare the rendered and observed scene is
based purely on 3D sensor data, as opposed to the 2D edge-feature and
per-pixel depth differences used in RMR that make it vulnerable to offset
errors between the rendered and observed 2D scenes. Moreover, the explanation
cost we propose can be decomposed over the objects in the scene, thereby
obviating the need for exhaustive search over the joint object poses.

Finally, an emerging trend for object recognition and pose estimation in RGB-D
data is the use of deep neural networks trained on synthetic
data generated using 3D models~\cite{schwarzrgb,wu20153d}. As promising as
deep learning methods are, they would require sufficient training data to capture
invariances to multi-object interaction and occlusion, the generation of which
is a combinatorial problem by itself. On the other hand, these
methods could be incorporated in PERCH as heuristics for guiding deliberative search as discussed in
Sec.~\ref{subsec:heuristics}.

\section{Problem Formulation}
\label{sec:formulation}

The problem we consider is that of localizing tabletop objects in a point
cloud or 2.5D data such as from a Kinect sensor. The problem
statement is as follows: given 3D models of $N$ unique objects, a point cloud
($I$) of a scene containing $K\geq N$ objects (possibly containing replicates of the $N$
unique objects), and the 6 DoF pose of the sensor, we are required to find
the 3 DoF pose ($x, y, \theta$) of each of the $K$ objects in the scene. 

We make the following assumptions:
\begin{itemize}
    \item The number ($K$) and type of objects in the scene are known ahead of
      time
      (but not the correspondences themselves).
    \item The objects in the scene vary only in position $(x,y)$ and yaw
      $(\theta)$---3 DoF, with respect
      to their 3D models.
    \item The input point cloud can be preprocessed (table plane, background
      filtered etc.) such that the points in it only belong to objects of
      interest.
    \item We have access to the intrinsic parameters of the sensor, so
      that we can render scenes using the available 3D models.
\end{itemize}

We specifically note that we \emph{do not} make any assumptions about the ability to
`cluster' points into different object groups as is done by most global 3D object
recognition methods such as the Viewpoint Feature Histogram (VFH)~\cite{rusu2010fast}.

\subsection{Notation}
Throughout the paper, we will use the following notation:

\begin{itemize}
  \item $O$: An object state characterized by $(\text{ID}, x, y, \theta)$, the
    unique object ID, position and yaw.
    \item $I$: The input/observed point cloud from the depth sensor.
    \item $R_j$: A point cloud generated by rendering a
      scene containing objects $O_1, O_2, \ldots, O_j$.
    \item $p$: A point $(x, y, z)$ in any point cloud.
    \item $\Delta R_j = R_j - R_{j-1}$, the point cloud containing points in $R_j$ but not in
      $R_{j-1}$. In other words, the set of points belonging to object $O_j$ that would
      be visible given the presence of objects $O_1, O_2,\ldots,O_{j-1}$.
    \item $V(O_j)$: The set of all points in the volume occupied by object
      $O_j$. When it is not possible to compute this in closed form, this can
      be replaced by an admissible/conservative approximation, for example,
      the volume of an inscribed cylinder.
    \item $V_j = \cup_{i=1}^j V(O_i)$, the union of volumes occupied by
      objects $O_1, O_2, \ldots, O_j$.
\end{itemize}

\subsection{Explanation Cost Function}

\begin{figure}[t]
  \centering
  \qquad
  \includegraphics[height=3cm]{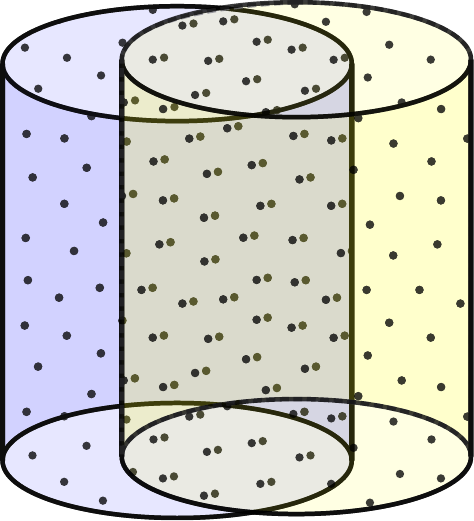}
  \includegraphics[height=3cm]{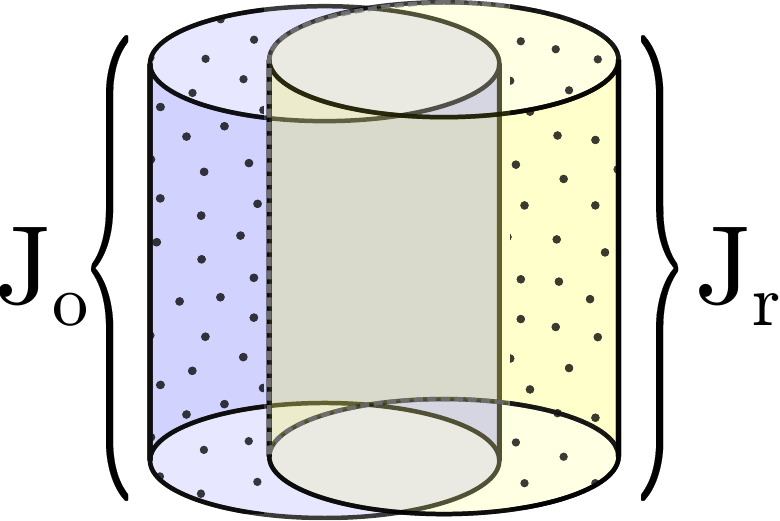}
  \caption{Illustration showing the computation of the explanation cost.
  The figure on the left shows the superposition of the observed point cloud
  (in blue) and the rendered point cloud (in yellow) of a cylindrical object.
  Object boundaries and volumes are shown merely for illustration. The total
  explanation cost (see figure on the right) is the number of unexplained
  points in the observed point cloud ($J_o$) and the number of unexplained
  points in the rendered point cloud ($J_r$).}
  \label{fig:cost}
  \vspace{-4mm}
\end{figure}

We formulate the problem of identifying and obtaining the 3 DoF poses of
objects $O_1,O_2, \ldots, O_K$ as that of finding the
minimizer of the following `explanation cost':
\begin{align*}
  J(O_{1:K}) &= J_{observed}(O_{1:K}) + J_{rendered}(O_{1:K}) \\
  J_{observed}(O_{1:K}) &= \sum_{p \in I} \mathds{1}_{[p \text{ is unexplained by }
  R_K]} \\
  J_{rendered}(O_{1:K}) &= \sum_{p \in R_K}\mathds{1}_{[p \text{ is unexplained by }
  I]}
\end{align*}

in which the indicator function $\mathds{1}_{[p \text{ is unexplained by }
C]}$ for a
point cloud $C$ and point $p$ is defined as follows:

\begin{align*}
  \mathds{1}_{[p \text{ is unexplained by } C]} &= \begin{cases} 1 & \text{if }
    \min_{p'
    \in C}\|p'-p\| > \delta \\
    0  & \text{otherwise} \numberthis \label{eq:explanation_indicator}
  \end{cases}
\end{align*}

for some sensor noise threshold $\delta$. We will use the notation $J_o$ and
$J_r$ to refer to the observed and rendered explanation costs respectively.

The explanation cost essentially counts the number of points in the observed
scene that are not explained by the rendered scene and the number of points in
the rendered scene that cannot be explained by the observed scene. While it
looks simplistic, the cost function forces the rendering of a scene that as
closely explains the observed scene as possible, from both a `filled'
(occupied) and `empty' (negative space) perspective. Figure~\ref{fig:cost}
illustrates the computation of the `explanation cost'. Another interpretation
for the explanation cost is to treat it as an approximation of the difference
between the union volume and intersection volume of the objects in the
observed and rendered scenes.

In the ideal scenario where there is no noise in the
observed scene and where we have access to a perfect renderer, we could do an
exhaustive search over the joint object poses to obtain a solution with zero
cost. However, this naive approach is a recipe for computational disaster:
even when we have only 3 objects in the scene and discretize our positions to
100 grid locations and 10 different orientations, we would have to
synthesize/render $10^9$ scenes to find the global optimum.  This immediately
calls for a better optimization scheme, which we derive next.

\section{PERCH: PErception via SeaRCH}
\label{sec:perch}
\subsection{Monotone Scene Generation Tree}
The crux of our algorithm exploits the insight that the explanation cost
function can be decomposed over the set of objects in the scene. To see this,
we first note that the rendered scene containing $K$ objects, $R_K$ can be
incrementally constructed:
\begin{align*}
  R_K &= \cup_{i=1}^K \Delta R_{i} &\text{s.t. } R_{i-1} \subseteq R_i
\end{align*}
where $\Delta R_i = R_i - R_{i-1}$ and $R_0$ is assumed to be an empty point cloud.
The constraint $R_{i-1} \subseteq R_i$ translates to saying that the addition of a
new object to the scene does not `occlude' the existing scene, thereby
guaranteeing that every point in $R_{i-1}$ exists in $R_{i}$ as well. In other
words, the number of points in the rendered point cloud can only increase with
the addition of a new object.
The above constraint implicity assumes that the scene does not contain objects
which can simultaneously occlude an object and also be occluded by another
object, such as horseshoe-shaped objects\footnote{In theory, we could still handle such objects 
by decomposing them into multiple surfaces
that satisfy the non-occlusion constraint. We omit the details for
simplicity of explanation.}. Using the above, we can write the rendered
explanation cost as follows:
\begin{align*}
  J_{r} &= \sum_{p \in R_K}\mathds{1}_{[p \text{ is unexplained by }
  I]} &\\
                      &= \sum_{i=1}^K\sum_{p \in \Delta
                      R_i}\mathds{1}_{[p \text{ is unexplained by }
                      I]} & \text{ s.t. } R_{i-1} \subseteq R_{i}
\end{align*}
We then similarly decompose the observed explanation cost:
\begin{align*}
  J_{o} &= \sum_{p \in I} \mathds{1}_{[p \text{ is unexplained by }
  R_K]} &\\
        &= \sum_{p \in I} \prod_{i=1}^K\mathds{1}_{[p \text{ is unexplained by }
  \Delta R_i]} & \text{ s.t. } R_{i-1} \subseteq R_{i}\\
               &= \sum_{i=1}^K\sum_{p \in \{I \cap V(O_i)\}} \mathds{1}_{[p
  \text{ is unexplained
by }
  \Delta R_i]}\\
                &\;\qquad+ \sum_{p \in \{I - V_K\}}
  \mathds{1}_{[p \text{ is unexplained by }
  R_K]} &\text{ s.t. } R_{i-1} \subseteq R_{i}
\end{align*}

With the above decompositions, we can re-write the overall optimization objective
as:
\begin{align*}
  J(O_{1:K}) &= \sum_{i=1}^K \Delta J^i  &\text{
  s.t. } R_{i-1} \subseteq R_{i} \numberthis\label{eq:total_cost}\\
  &= \sum_{i=1}^K \Delta J_{r}^i + \Delta J_{o}^i &\text{
  s.t. } R_{i-1} \subseteq R_{i}
\end{align*}
where 
\begin{align*}
  \Delta J_{r}^i &= \sum_{p \in \Delta
R_i}\mathds{1}_{[p \text{ is unexplained by }
I]} \\
  \Delta J_{o}^i &= 
\sum_{p \in \{I \cap V(O_i)\}} \mathds{1}_{[p \text{ is unexplained by }
\Delta R_i]} + \text{residual}(i)\\
  \text{residual}(i) &= \begin{cases}
              \sum_{p \in \{I - V_K\}}
              \mathds{1}_{[p \text{ is unexplained by }
              R_K]} &\text{ if } i = K\\
    0 &\text{ otherwise }
  \end{cases}
\end{align*}

\begin{figure}[t]
  \centering
  \includegraphics[width=\columnwidth]{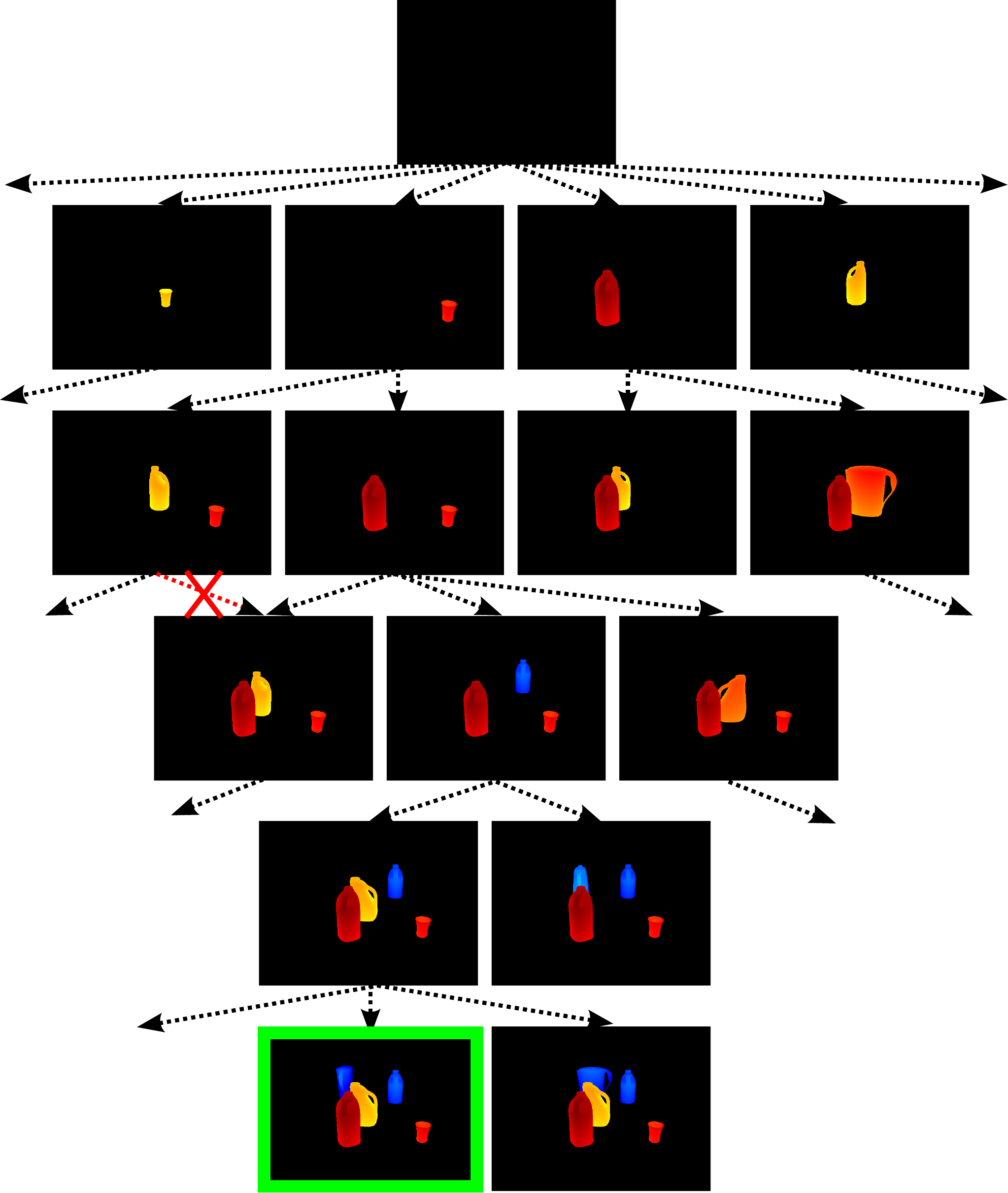}
  \caption{Portion of a Monotone Scene Generation Tree (MSGT): the root
    of the tree is the empty scene, and new objects are added progressively as
    we traverse down the tree. Notice how child states never introduce an
    object that occludes objects already in the parent state. A
    counter-example (marked by the red cross) is also shown. Any state on the
    $K^{\text{th}}$ level of the tree is a goal state, and the task is to find
  the one that has the lowest cost path from the root---marked by a green
  bounding box in this example.}
  \label{fig:msgt}
  \vspace{-4mm}
\end{figure}
\begin{figure*}[th!]
  \centering
  \includegraphics[width=\linewidth]{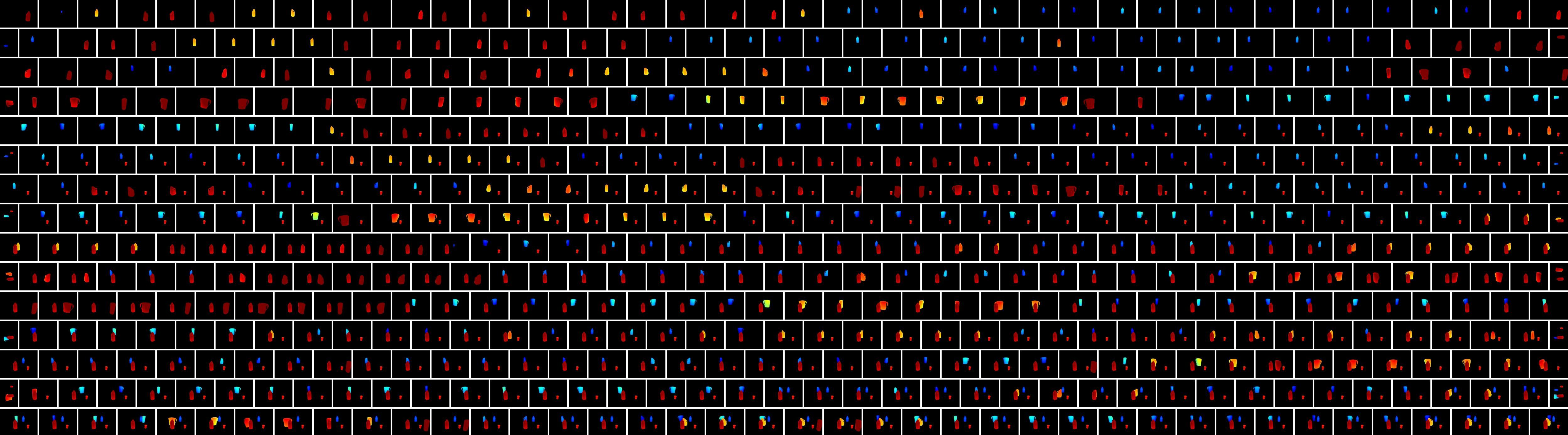}
  \caption{A subset of all the states `generated' during the tree search for
    the scene in Fig.~\ref{fig:msgt}. This figure is best viewed with digital
  zoom.}
  \label{fig:mosaic}
\vspace{-3mm}
\end{figure*}

Equation~\ref{eq:total_cost} defines a pairwise-constrained optimization problem,
the constraint being that the assignment of the $i^\text{th}$ object does not
occlude the scene generated by the assignment of the previous objects $1$
through $i-1$. A natural way to solve this problem is to construct a tree that
satisifies the required constraint, and assigns the object poses in a sequential
order. This is precisely our approach and the resulting tree we construct is
called the Monotone Scene Generation Tree (MSGT), with `monotone' emphasizing that as we go down the
tree, newly added objects cannot occlude the scene generated thus
far (Fig.~\ref{fig:msgt}). We note
that while a particular configuration of objects can be generated
by choosing different assignment orders, only one is sufficient to retain as
all those configurations have identical explanation costs. Thus, we obtain a tree structure
instead of a Directed Acyclic Graph (DAG).
Formally, any vertex/state in the MSGT is a partial assignment of object states: $s
= \{O_{1:j}\}$, with
$j\leq K$. For a MSGT state $s$ with an assignment of $j$ objects, the
implicit successor generation function and the associated cost are defined as
follows:
\begin{align*}
  \textsc{Succ}(s) &= \{s' | s' = s \cup O_{j+1} \land R_{j} \subseteq
  R_{j+1}\} \numberthis \label{eq:succ}\\
  \textsc{c}(s, s') &= \Delta J^{j+1} = \Delta J_r^{j+1} + \Delta J_o^{j+1} \numberthis \label{eq:cost}
\end{align*}
The root node of the tree $s_{\text{root}}$ is an empty state containing no
object assignments, while a goal state is any state $s$ that has an assignment
for all objects. Given the MSGT construction, the multi-object localization problem reduces to
that of finding the cheapest cost path in the tree from the root state to any goal state.

\subsection{Tree Search}
Although we have replaced exhaustive search with tree search, the
problem still remains daunting owing to its branching factor. Assume that we
have $d$ possible configurations $(x, y, \theta)$ for each object. Then, the
worst case branching factor for the MSGT is $d^{K}$ for all levels if we allow
repetition of objects in the scene, or $d^{K-i}$ for level $i$ if there is no
repetition. Figure~\ref{fig:mosaic} illustrates this by showing a subset of
the states generated during the tree search corresponding to the scene in
Fig.~\ref{fig:msgt}. While heuristic search techniques such as A* are often a good
choice for such problems, they require an \emph{admissible} heuristic that
provides a conservative estimate of the remaining cost-to-go. Usual heuristic
search methods are limited by the following: i) admissible heuristics are
non-trivial to obtain for this problem, and ii) they cannot support multiple
heuristics, each of which could be useful on their own---for e.g, different
feature-based and learning-based methods could serve as a heuristic each.
Fortunately, recent work in heuristic search~\cite{narayanan2015improved,
aine2014multi} allows us to use multiple, inadmissible heuristics to find
solutions with bounded suboptimality guarantees. 

The particular multi-heuristic search we use is the
Focal-MHA*~\cite{narayanan2015improved} algorithm, and its choice is motivated
by the fact that it permits the use of inadmissible heuristics that have no
connection with the cost structure of the problem. This necessity will become
clear in Sec.~\ref{subsec:heuristics}. At a high level, Focal-MHA* operates
much like A* search. Like A*, it maintains a priority list of
states ordered by an estimate of the path cost through that state and
repeatedly `expands' the most promising states until a goal is found. The
difference from A* is in that Focal-MHA* interleaves this process with
expansion of states chosen greedily by other
heuristics~\cite{narayanan2015improved}. Finally, Focal-MHA* guarantees that
the solution found will have a cost which is bounded by $w\cdot \text{OPT}$,
where OPT is the optimal solution cost and $w(\geq 1)$ is a user-chosen
suboptimality bound. Algorithm~\ref{alg:perch} shows
an instantiation of Focal-MHA* in the context of PERCH.

\begin{algorithm}[t!]
  \caption{PERCH}
  \label{alg:perch}
  \begin{algorithmic}
    \small
    \INPUT
    \State The implicit MSGT construction (Eq.~\ref{eq:succ} and Eq.~\ref{eq:cost}).
    \State Suboptimality bound factor $w$ ($\geq 1$).
    \State $1$ admissible heuristic $h$ and $n$ arbitrary, possibly
    inadmissible, heuristics $h_1, h_2,\ldots,h_n$.
    \\\hrulefill

    \OUTPUT
    \State An assignment of object poses $s_{\text{goal}}$ with $|s_{\text{goal}}|=K$ whose cost is within $w
    \cdot\text{OPT}$.
    \\\hrulefill
  \end{algorithmic}

  \begin{algorithmic}[1]
    \small

    \Procedure{Main()}{}
    \State $s_{\text{root}} \leftarrow \{\}$
    \State planner $\leftarrow$ Focal-MHA*-Planner()
    \State planner.$\textsc{SetImplicitTree}(\textsc{Succ}(s), c(s,s'))$
    \State planner.$\textsc{SetSuboptimalityFactor}(w)$
    \State planner.$\textsc{SetStartState}(s_{\text{root}})$
    \State planner.$\textsc{SetHeuristics}(h, h_1,\ldots,h_n)$
    \State planner.$\textsc{SetGoalCondition}(\text{\bf{return true} if }|s|=K)$
    \State $\{s_{\text{root}}, s_1, s_2,\ldots,s_{\text{goal}}\} \leftarrow$ planner.$\textsc{ComputePath}()$
    \State \Return $s_{\text{goal}}$
    \EndProcedure
  \end{algorithmic}
\end{algorithm}

\subsection{Heuristics}
\label{subsec:heuristics}
Focal-MHA* requires one admissible and multiple inadmissible heuristics.
Constructing an informative admissible heuristic is non-trivial for this
setting, and thus we set our admissible heuristic to the trivial heuristic
that returns $0$ for all states. We next describe our inadmissible heuristics.

The large branching factor of the MSGT might result in the search `expanding' or
opening every node in a level before moving on to the next. To guide the
search towards the goal, a natural heuristic would be a depth-first heuristic that
encourages expansion of states further down in the tree. Consquently, our
first inadmissible heuristic for Focal-MHA* is the depth heuristic that
returns the number of assignments left to make:
\begin{align*}
  h_{depth}(s) =  K - |s|
\end{align*}

As a reminder, states with \emph{smaller} heuristic values are expanded ahead
of those with larger values. Next, it would be useful to encourage the search to expand states that have
maximum overlap with the observed point cloud $I$ so far, rather than states with
little overlap with the observed scene. Our second heuristic is therefore the
overlap heuristic that counts the number of points in $I$ that \emph{do not} fall within the
volume of assigned objects:
\begin{align*}
  h_{overlap}(s) = |I|-\sum_{p \in I} \mathds{1}_{[p \in V_j]}
\end{align*}
where $j = |s|$ and $V_j$ is the union of
the volumes occupied by the $j$ assigned objects. Another interpretation for this
heuristic is the number of points in the observed scene that are outside the
space carved by objects assigned thus far.

While we use only the above two heuristics in this work, we note that there is
a possibility of using a wide class of heuristics derived from feature and
learning-based methods. For instance, if an algorithm like
VFH~\cite{rusu2010fast} produced a pose estimate $O_j', j=1,2,\ldots, K$ for
each of the $K$ objects in the scene, then a heuristic for $s$ with $|s|=j$
could resemble $\sum_{i=1}^j \|O_i-O_i'\|$ for some appropriate choice of the
norm. More generally, the multi-heuristic approach we use provides a framework
to plug in various discriminative algorithms each with their own strengths and weaknesses.

\subsection{Theoretical Properties}
PERCH inherits all the theoretical properties of Focal-MHA*~\cite{narayanan2015improved}.
We state those here without proof:
\begin{theorem}
PERCH is complete with respect to the
construction of the graph, i.e, if a solution (feasible assignment of all object
poses) exists in the MSGT, it will be found. 
\end{theorem}
\begin{theorem}
The returned solution has an explanation cost which is no more than $w$ times
the cost of the best possible solution under the chosen graph construction. 
\end{theorem}
As a disclaimer, we note that bounded suboptimal solutions with regard to the
explanation cost do not translate to any bounded suboptimal properties with
respect to the true object poses in the observed scene.

\subsection{Implementation Details}
\subsubsection{Compensating for Discretization}
\label{subsubsec:discretization}
The most computationally complex part of PERCH is that of generating
successor states for a given state in the MSGT. This involves generating and
rendering every state that contains one more object than the number in the
present state, in every possible configuration. Several elements influence
this branching factor: the number of objects in the scene, the chosen
discretization for object poses, whether objects are rotationally symmetric
(in which case only $(x,y)$ is of interest) etc. In our implementation, we
limit the complexity by favoring coarse discretization and compensating with
a local alignment technique such as ICP~\cite{chen1991object}. Specifically,
every time we render a state with a new object, we take the non-occluded
portion and perform an ICP alignment in the local vicinity of the observed
point cloud. This allows us to obtain accurate pose estimates while retaining
a coarse discretization. We do note that the underlying MSGT now becomes a
function of the observed point cloud due to the ICP adjustment.

\subsubsection{Parallelization}
The generation of successor states is an embarassingly parallel process. We
exploit this in our implementation by using multiple processes to generate
successors in parallel. Theoretically, with sufficient number of cores, the time to expand a
state would simply be the time to render a single scene.

\section{Experiments}
\label{sec:experiments}
\subsection{Occlusion Dataset}
\begin{figure}[t]
  \centering
  \begin{subfigure}[t]{0.4\columnwidth}
  \includegraphics[width=1\columnwidth]{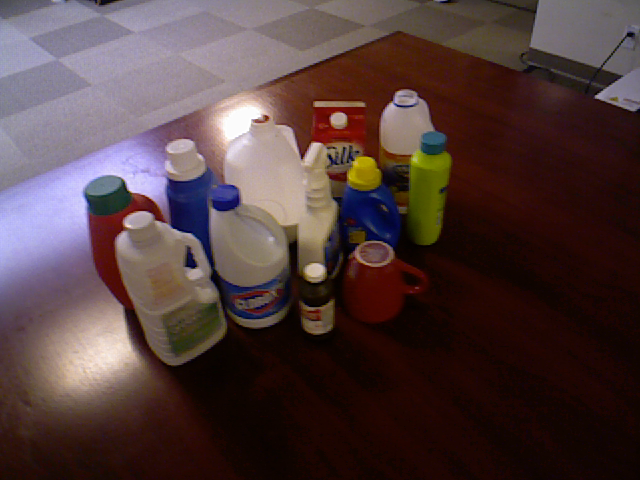}
  \caption{}
  \label{dataset_objects}
  \end{subfigure}
  \begin{subfigure}[t]{0.4\columnwidth}
  \includegraphics[width=1\columnwidth]{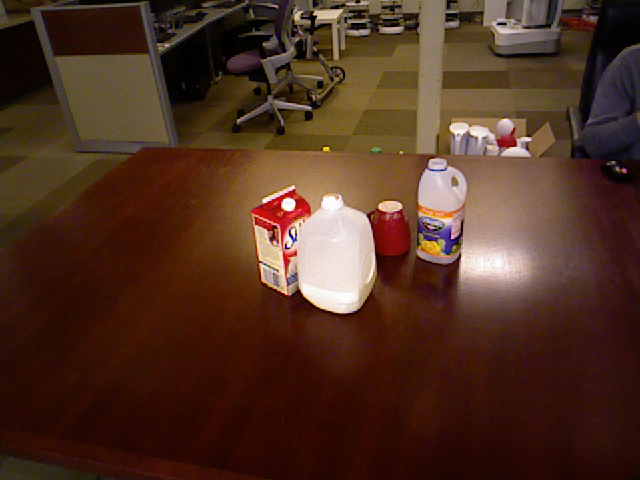}
  \caption{}
  \label{dataset_scene}
  \end{subfigure}\\
  \vspace{1mm}
  \begin{subfigure}[t]{0.4\columnwidth}
  \includegraphics[width=1\columnwidth]{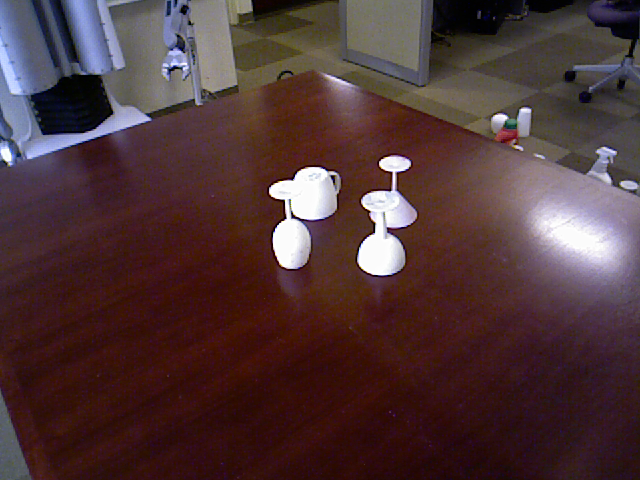}
  \caption{}
  \label{dataset_rgb}
  \end{subfigure}
  \begin{subfigure}[t]{0.4\columnwidth}
  \includegraphics[width=1\columnwidth]{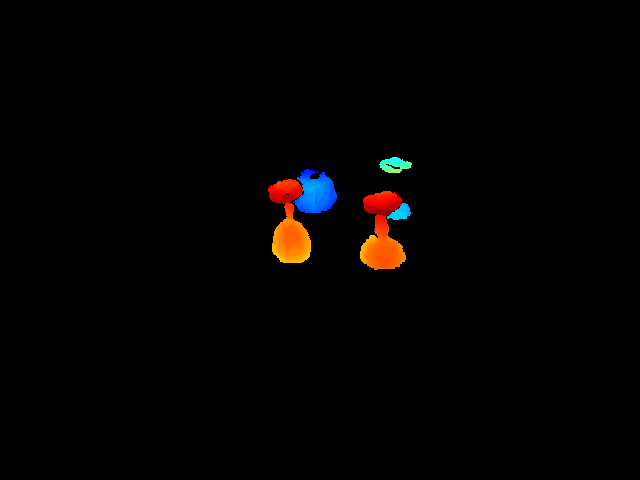}
  \caption{}
  \label{dataset_depth}
  \end{subfigure}
  \caption{\subref{dataset_objects} A subset of the objects in the RGB-D
    occlusion dataset.
  \subref{dataset_scene} An example scene from the dataset. 
  \subref{dataset_rgb} Another example from the dataset and
  \subref{dataset_depth} the corresponding depth image (mapped to jet color) obtained after
  preprocessing. In this example, portion of the occluded glass is missing due to
  sensor noise, making it challenging for methods that rely on segmenting
  the scene into cohesive object point clouds.}
  \label{fig:dataset}
\vspace{-4mm}
\end{figure}
\begin{figure*}[th!]
  \centering
  \begin{subfigure}[t]{0.325\textwidth}
  \includegraphics[width=\columnwidth]{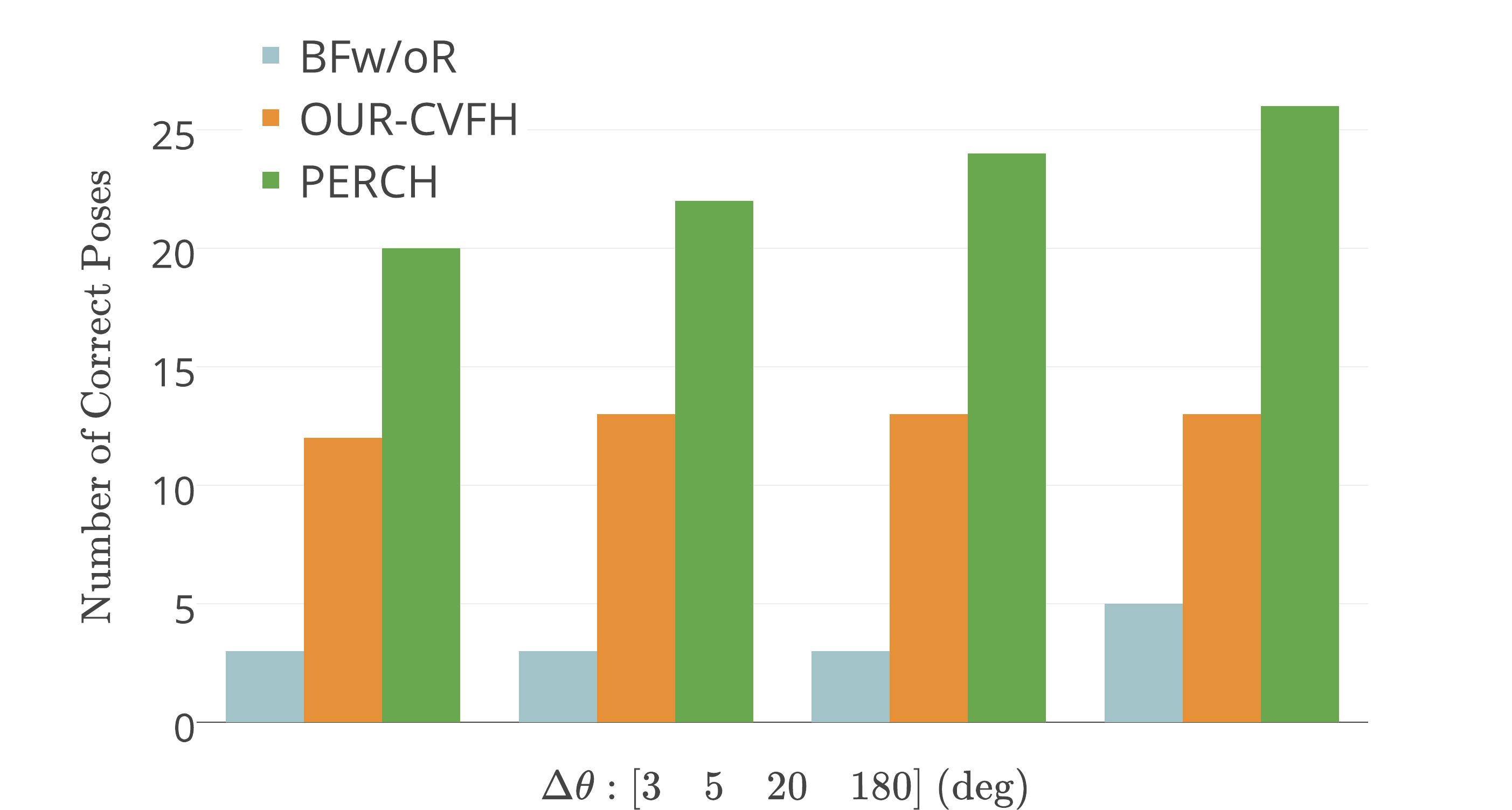}
  \caption{$\Delta t = 0.01m$}
  \label{fig:first_hist}
  \end{subfigure}
  \begin{subfigure}[t]{0.325\textwidth}
  \includegraphics[width=\columnwidth]{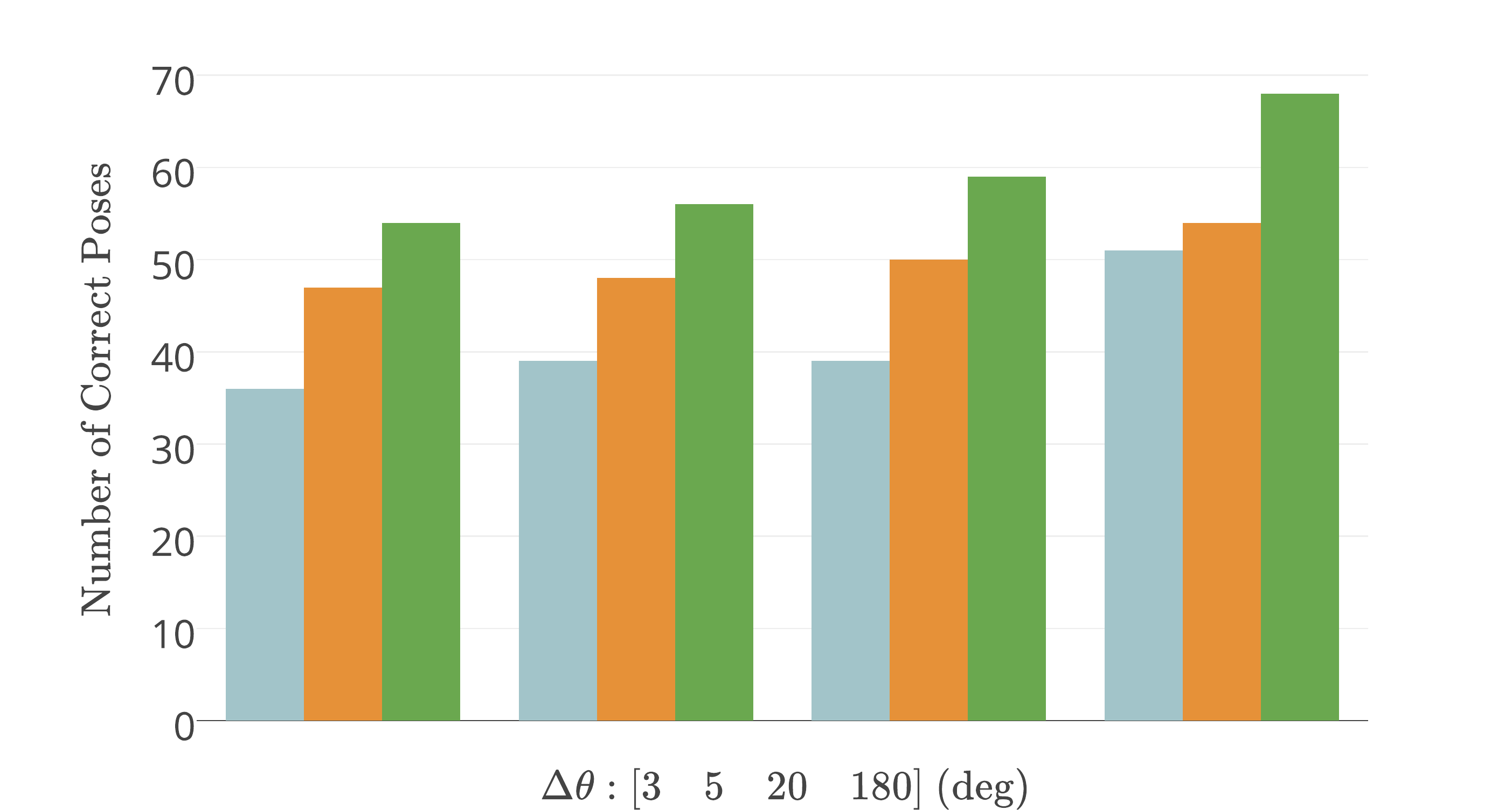}
  \caption{$\Delta t = 0.05m$}
  \label{fig:second_hist}
  \end{subfigure}
  \begin{subfigure}[t]{0.325\textwidth}
  \includegraphics[width=\columnwidth]{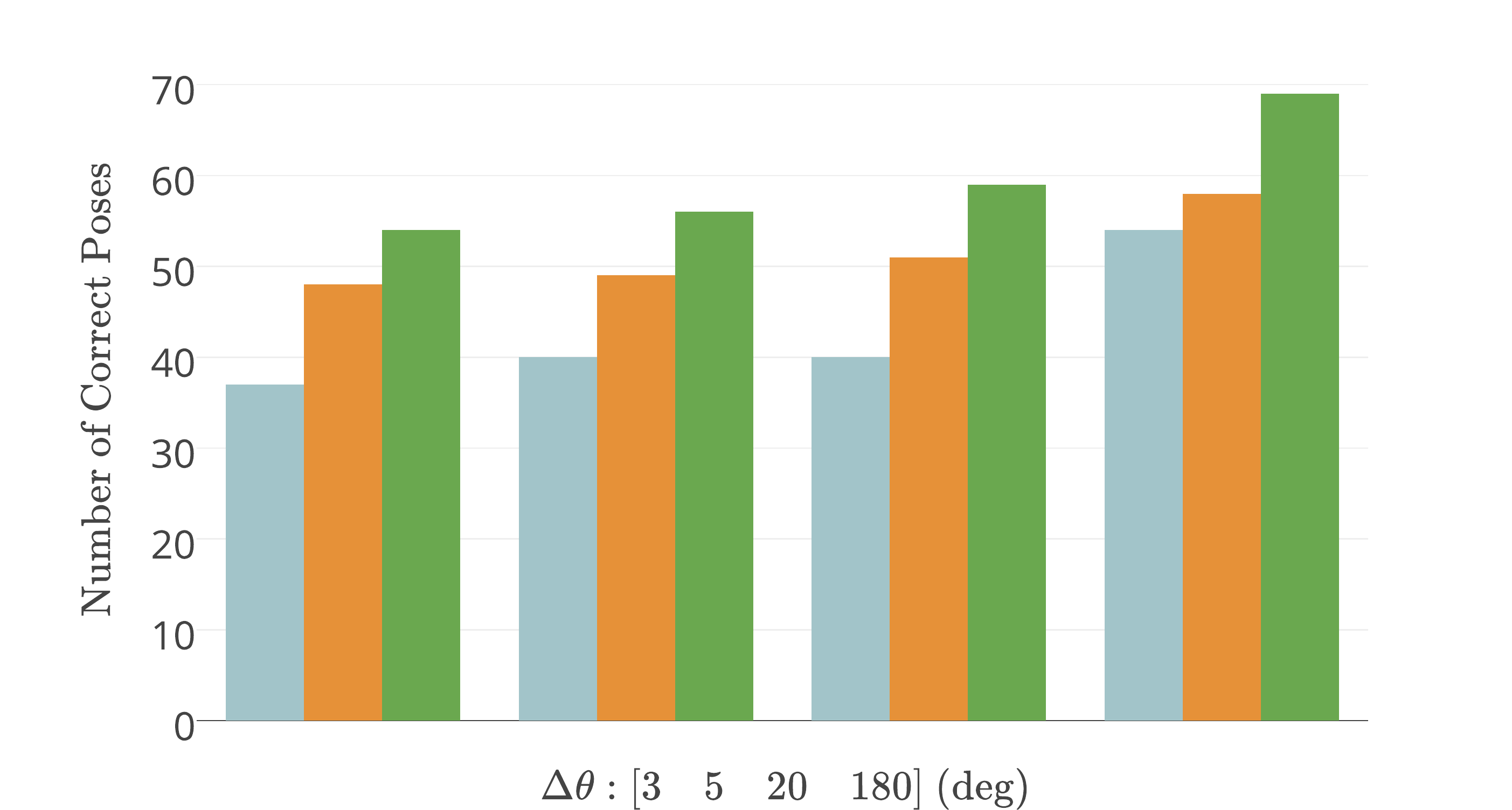}
  \caption{$\Delta t = 0.1m$}
  \label{fig:last_hist}
  \end{subfigure}
  \caption{Number of objects whose poses were correctly classified by the
  baseline methods (BFw/oR, OUR-CVFH) and PERCH, for different definitions of
`correct pose'.}
  \label{fig:experiments}
\vspace{-4mm}
\end{figure*}

To evaluate the performance of PERCH for multi-object recognition and pose
estimation in challenging scenarios where objects could be occluding each
other, we pick the occlusion dataset described by Aldoma et
al.~\cite{aldoma2012point} that contains objects partially touching and
occluding each other. The dataset contains 3D CAD models of 36 common
household objects, and 23 RGB-D tabletop scenes with 82 object instances in total.
All scenes except one contain objects only varying in translation and yaw,
with some objects flipped up-side down. Since PERCH is designed only for 3D pose
estimation, we drop the one non-compatible scene from the dataset, and
preprocess the 3D CAD models such that they vary only in translation and yaw
with respect to the ground truth poses. Figure~\ref{fig:dataset} shows some
examples from the dataset.

\subsection{PERCH Setup}
Since PERCH requires that points in the scene only belong to objects of
interest, we first preprocess the scene to remove the tabletop and background.
Then, based on the RANSAC-estimated table plane, we compute a transform that
aligns the point cloud from the camera frame to a gravity aligned frame, to
simplify construction of the MSGT. PERCH has two parameters to set:
the sensor noise threshold $\delta$ for determining whether a point is
`explained' (Eq.~\ref{eq:explanation_indicator}), and the suboptimality factor $w$
for the Focal-MHA* algorithm. In our experiments, we set $\delta$ to $3$ mm to
account for uncertainty in the depth measurement from the Kinect sensor, as
well as inaccuracies in estimating the table height using RANSAC. For the
suboptimality factor $w$, we use a value of $3$. While this results in
solutions that can be suboptimal by a factor of upto $3$, it greatly speeds
up the search since computing the optimal solution typically takes much more
time~\cite{pohl70}. For Focal-MHA*, we use the two heuristics described in Sec.~\ref{subsec:heuristics}.
Finally, for defining the MSGT we pick a discretization resolution of $4$ cm
for both $x$ and $y$ and $22.5$ degrees for yaw.
The adaptive ICP alignment (Sec.~\ref{subsubsec:discretization}) is
constrained to find correspondences within $2$ cm, which is half the discretization resolution.

\subsection{Baselines}
Our first baseline is the OUR-CVFH global descriptor~\cite{aldoma2012our}, a
state-of-the-art global descriptor designed to be robust to occlusions. By
clustering object surfaces into separate smooth regions and computing
descriptors for each portion, OUR-CVFH can handle occlusions better than
descriptors such VFH and FPFH. Furthermore, it has the added advantage of
directly encoding the full pose of the object, with no ambiguity in camera
roll. We build the training database by rendering 642 views of every 3D CAD
model from viewpoints sampled around the object. Then, for computing the
training descriptors we use moving least squares to upsample every training
view to a common resolution followed by downsampling to the Kinect resolution
of $3$ mm as suggested in the OUR-CVFH paper~\cite{aldoma2012our}. Since the
number and type of models in the test scene is assumed to be known for PERCH,
we use the following pipeline for fair comparison: for the $K$ largest
clusters in the test scene we obtain the histogram distance to each of the
models we know that are in the scene. Then, we solve a min-cost matching
problem to assign a particular model (and associated pose) to each cluster and
obtain a feasible solution. Finally, we constrain the full 6 DoF poses returned
by OUR-CVFH to vary only in translation and yaw and perform a local ICP
alignment for each object pose.

The second baseline is an ICP-based optimization one, which we will
refer to as Brute Force without Rendering (BFw/oR). Here, we slide the 3D model
of every object in the
scene over the observed point cloud (at the same discretization used for
PERCH), and perform a local ICP-alignment at every step. The location
($x,y,\theta)$ that has
the best ICP fitness score is chosen as the final pose for that model and
made unavailable for other objects that have not yet been considered. Since the order in
which the models are chosen for sliding can influence the solution, we try all permutations of the ordering ($K!$)
and take the overall best solution based on the total ICP fitness score. 

\subsection{Evaluation}

\begin{figure}[t]
  \centering
  \begin{subfigure}[t]{0.155\textwidth}
  \includegraphics[width=1\columnwidth]{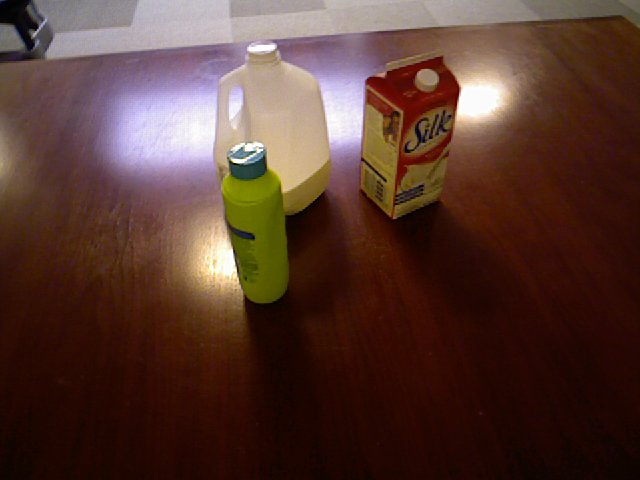}
  \end{subfigure}
  \begin{subfigure}[t]{0.155\textwidth}
  \includegraphics[width=1\columnwidth]{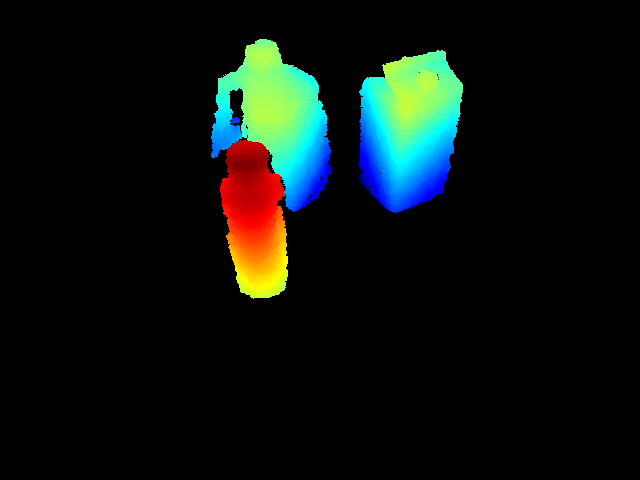}
  \end{subfigure}
  \begin{subfigure}[t]{0.155\textwidth}
  \includegraphics[width=1\columnwidth]{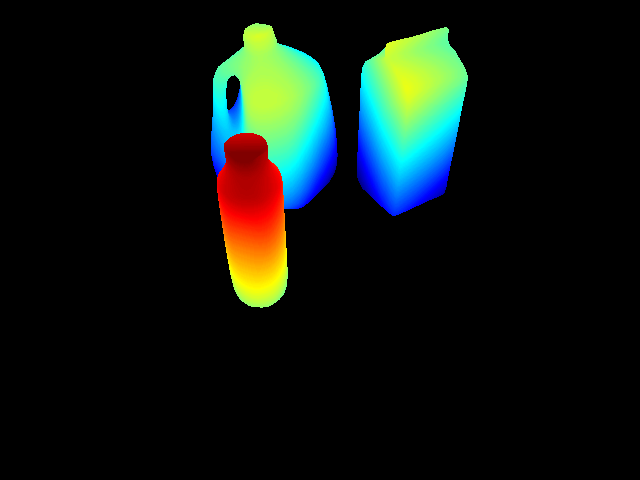}
\end{subfigure}\vspace{1mm}
  \begin{subfigure}[t]{0.155\textwidth}
  \includegraphics[width=1\columnwidth]{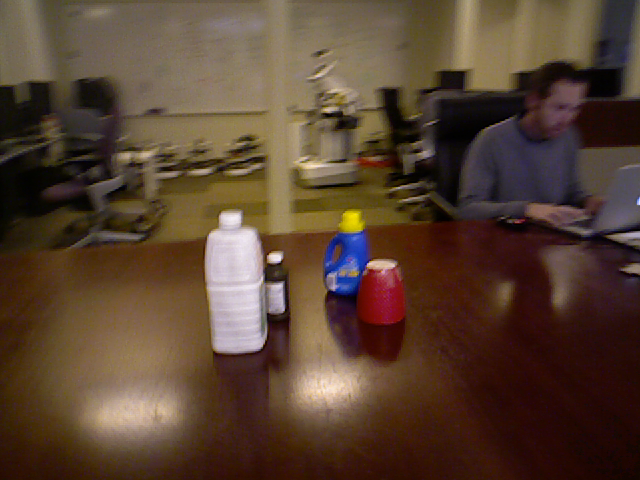}
  \end{subfigure}
  \begin{subfigure}[t]{0.155\textwidth}
  \includegraphics[width=1\columnwidth]{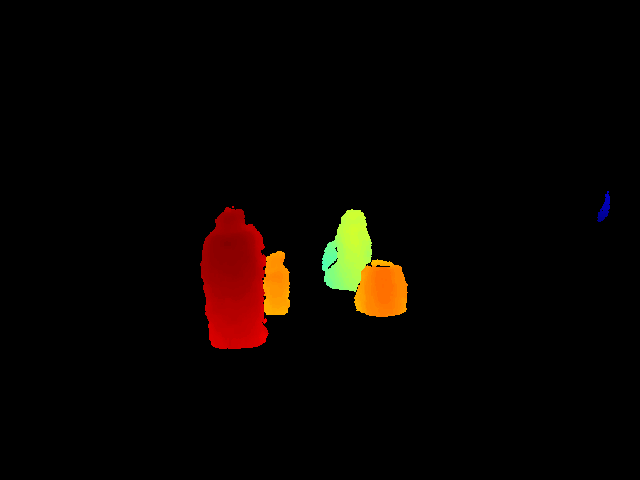}
  \end{subfigure}
  \begin{subfigure}[t]{0.155\textwidth}
  \includegraphics[width=1\columnwidth]{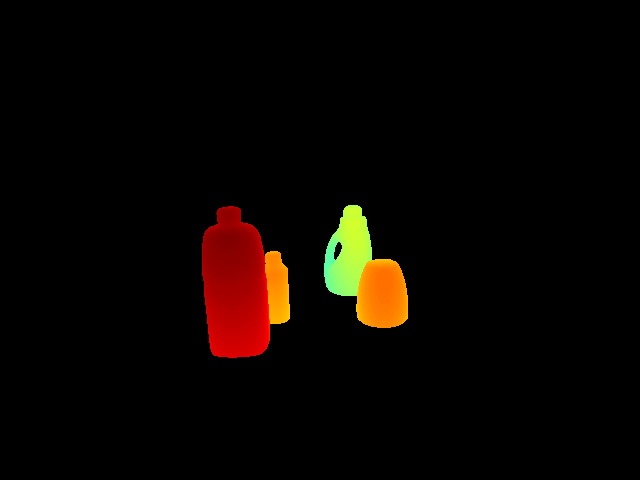}
\end{subfigure}\vspace{1mm}
  \begin{subfigure}[t]{0.155\textwidth}
  \includegraphics[width=1\columnwidth]{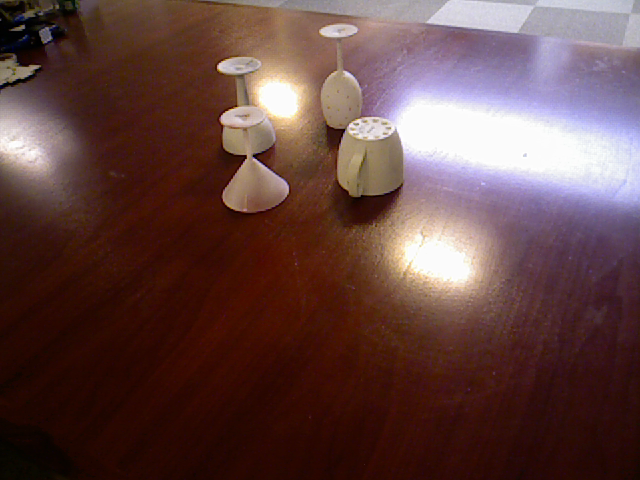}
  \end{subfigure}
  \begin{subfigure}[t]{0.155\textwidth}
  \includegraphics[width=1\columnwidth]{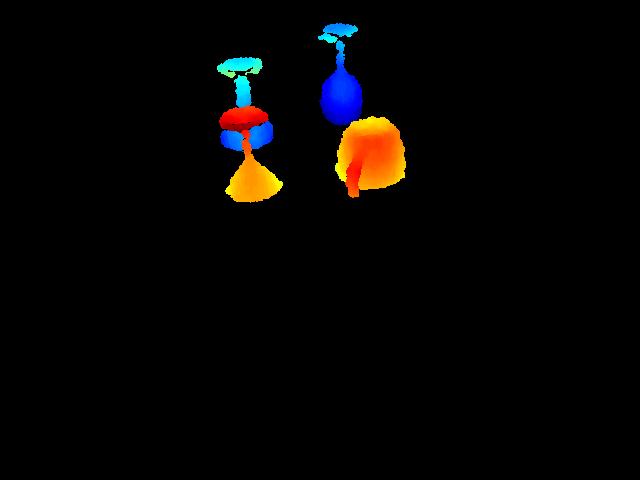}
  \end{subfigure}
  \begin{subfigure}[t]{0.155\textwidth}
  \includegraphics[width=1\columnwidth]{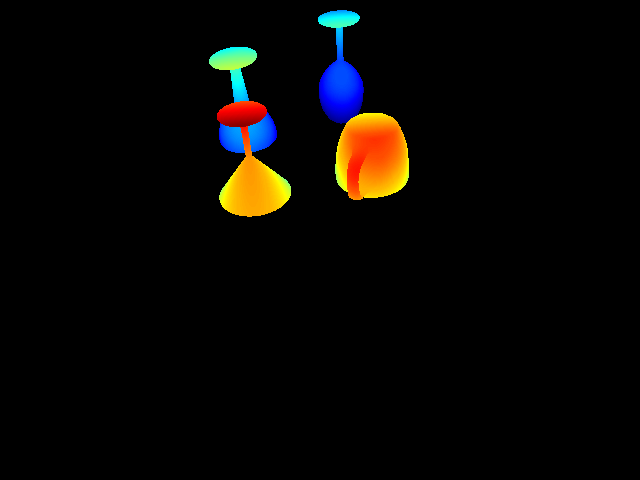}
  \end{subfigure}
  \caption{Examples showing the output of PERCH on the occlusion dataset.
    \emph{Left}: RGB-D scenes in the dataset. 
    \emph{Middle}: Depth images of
    the corresponding input RGB-D scenes,
    \emph{Right}: The depth image
  reconstructed by PERCH through rendering object poses.}
  \label{fig:experiment_examples}
\vspace{-4mm}
\end{figure}

To evaluate the accuracy of object localization, we use the following
criterion: a predicted pose $(x,y,\theta)$ for an object is considered correct if
$\|(x,y)-(x_{\text{true}},y_{\text{true}})\|_2 < \Delta t$ and
$\textsc{ShortestAngularDifference}(\theta,\theta_{\text{true}}) < \Delta \theta$. We
then compute the number of correct poses produced by each method for different
combinations of $\Delta t$ and $\Delta \theta$. Figure~\ref{fig:experiments} compares the performance of PERCH with BFw/oR and
OUR-CVFH. Immediately obvious is the significant performance
of PERCH over the baseline methods for $\Delta t=0.01m$. PERCH is able to
correctly estimate the pose of over 20 objects with translation error under
$1$ cm and rotation error under $5$ degrees. While the baseline methods have
comparable recall for higher thresholds, they are unable to provide as many
precise poses as PERCH does.  Further, PERCH consistently dominates the
baseline methods for all definition of `correct pose'. Among all methods,
BFw/oR performs the worst. This is mainly due to the fact that it uses the
point cloud corresponding to the complete object model for ICP refinement,
rather than the point cloud corresponding to the unoccluded portion of the
object. Again, this showcases the necessity to explicity reason about
self-occlusions as well as inter-object occlusions. 

The last column of the histogram in Fig.~\ref{fig:last_hist} (corresponding to
$\Delta t = 0.1$, $\Delta \theta = 180$) is essentially a measure of
recognition alone---PERCH can correctly identify 69 of the 80 object
instances, where `identified' is defined as obtaining a translation error
under $10$ cm.  Figure~\ref{fig:experiment_examples} shows some qualitative
examples of PERCH's peformance on the occlusion dataset. Further examples and
illustrations are provided in the supplementary video.

\subsection{Computation Time and Scalability}
Unlike global descriptor approaches such as OUR-CVFH which require an
elaborate training phase to build a histogram library, PERCH does not require
any precomputation. Consequently, the run time cost is high owing to the
numerous scenes that need to be rendered. However, as mentioned earlier, the
parallel nature of the problem and the easy availability of cluster computing
makes this less daunting. For our experiments, we used the MPI framework to
parallelize the implementation and ran the tests on a cluster of 2 Amazon AWS
m4.10x machines, each having a 40-core virtual CPU. For each scene, we
used a maximum time limit of 15 minutes and took the best solution obtained
within that time. Overall, the mean planning time was 6.5 minutes, and the
mean number of hypotheses rendered (i.e, states generated) was
15564. 

Finally, to demonstrate that PERCH can be used for scenes containing several
objects, we conducted a test on a chessboard scene (Fig.~\ref{chess_real}). We
captured a Kinect depth image of the scene containing 12 pieces, of which 6
are unique and 4 are rotationally symmetric. We ran PERCH with suboptimality
bound factor $w=15$ and sensor resolution $\delta=7.5$ mm, and took the
best solution found within a time limit of 20 minutes. The solution found
(i.e., the depth image corresponding to the goal state) is shown in
Fig.~\ref{chess_depth}.

\section{Conclusions}
In his lecture on computer heuristics in 1985~\cite{feynman1985hypothesis},
Richard Feynman notes that if one had access to all the generative parameters
of a scene (lighting, model etc.), one could possibly generate every single
scene and take the best match to the observed data. We presented PERCH as a
first step towards this deliberative reasoning. The key contributions were the
formulation of multi-object recognition and localization as an optimization
problem and designing an efficient combinatorial search algorithm for the
same. We demonstrated how PERCH can robustly localize objects under occlusion,
and handle scenes containing several objects.

While our results look promising on the accuracy front, much work remains to
be done in making the algorithm suitable for real-time use. Our future work
involves exploring optimizations and heuristics for the search to obtain
faster yet high quality solutions. Specifically, we are interested in
leveraging state-of-the-art discriminative learning to provide guidance for
the search.  Other directions include generalizing PERCH to a variety of
perception tasks that require deliberative reasoning.

\addtolength{\textheight}{-2cm}   

\section*{Acknowledgment}
This research was sponsored by ARL, under the Robotics CTA program grant
W911NF-10-2-0016. We thank Maurice Fallon and Hordur Johannson
for making their Kinect simulator publicly available as part of the Point Cloud
Library.

\bibliographystyle{IEEEtranN}
\scriptsize{
\bibliography{perch}
}

\end{document}